\documentclass[final,5p]{elsarticle}

\usepackage[hidelinks]{hyperref}
\usepackage{lineno}
\usepackage{lipsum}
\usepackage{amsmath}
\usepackage{amssymb}
\usepackage{tabularx}
\usepackage{fancyhdr}
\usepackage{multirow}

\journal{arxiv.org}

\begin{document}
%%%%%%%%%%%%%%%%%%%%%%%%%%%%%%%%%%%%%%%%%%%%%%%%%%%%
\begin{frontmatter}

\title{Logical Learning Through a Hybrid Neural Network with Auxiliary Inputs}
\tnotetext[mytitlenote]{Original research contribution by \href{http://ancorasir.com}{the Sustainable + Intelligent Robotics Group}.}

\author[a]{WAN Fang}
\author[b]{SONG Chaoyang\corref{cor1}}

\address[a]{Independent Research Consultant, sophie.fwan@gmail.com}
\address[b]{Monash University, songcy@ieee.org}

\cortext[cor1]{Corresponding author.}
%%%%%%%%%%%%%%%%%%%%%%%%%%%%%%%%%%%%%%%%%%%%%%%%%%%%
\begin{abstract}
%%%%%%%%%%%%%%%%%%%%%%%%%%%%%%%%%%%%%%%%%%%%%%%%%%%%
The human reasoning process is seldom a one-way process from an input leading to an output. Instead, it often involves a systematic deduction by ruling out other possible outcomes as a self-checking mechanism. In this paper, we describe the design of a hybrid neural network for logical learning that is similar to the human reasoning through the introduction of an auxiliary input, namely the indicators, that act as the hints to suggest logical outcomes. We generate these indicators by digging into the hidden information buried underneath the original training data for direct or indirect suggestions. We used the MNIST data to demonstrate the design and use of these indicators in a convolutional neural network. We trained a series of such hybrid neural networks with variations of the indicators. Our results show that these hybrid neural networks are very robust in generating logical outcomes with inherently higher prediction accuracy than the direct use of the original input and output in apparent models. Such improved predictability with reassured logical confidence is obtained through the exhaustion of all possible indicators to rule out all illogical outcomes, which is not available in the apparent models. Our logical learning process can effectively cope with the unknown unknowns using a full exploitation of all existing knowledge available for learning. The design and implementation of the hints, namely the indicators, become an essential part of artificial intelligence for logical learning. We also introduce an ongoing application setup for this hybrid neural network in an autonomous grasping robot, namely \textit{as\_DeepClaw}, aiming at learning an optimized grasping pose through logical learning. 
\end{abstract}
%%%%%%%%%%%%%%%%%%%%%%%%%%%%%%%%%%%%%%%%%%%%%%%%%%%%
\begin{keyword}
%%%%%%%%%%%%%%%%%%%%%%%%%%%%%%%%%%%%%%%%%%%%%%%%%%%%
logic reasoning\sep hybrid neural network\sep deep learning\sep robotic grasping
\end{keyword}

\end{frontmatter}
% \linenumbers

\pagebreak
\tableofcontents

%%%%%%%%%%%%%%%%%%%%%%%%%%%%%%%%%%%%%%%%%%%%%%%%%%%%
\section{Introduction}
%%%%%%%%%%%%%%%%%%%%%%%%%%%%%%%%%%%%%%%%%%%%%%%%%%%%
Dictionary is a common tool that helps the humans to learn a native or new language. One key design principal of the dictionary is the use of \textit{circular definitions} to explain the meanings of more advanced or complex words by using a restricted list of high-frequency words, also known as the \textit{defining vocabulary}. For example, popular dictionaries such as the Longman Dictionary of Contemporary English, and the Oxford Advanced Learner's Dictionary contain defining vocabularies of a similar size with approximately 2,000-3,000 words. Essentially, such defining vocabulary becomes the core \textit{hints} that \textit{directly} or \textit{indirectly} suggest the meanings of more advanced vocabulary in the rest of these dictionaries. Moreover, the knowledge of the language is essentially embedded underneath such expressive process of suggestive explanation and exemplified usage. Once understood the meaning of a word, besides using it in daily lives, language learners usually adopt a revered process to reinforce their learning outcomes by accurately reciting the new word in pronunciation, spelling, and full meaning when supplied with some hints that indirectly suggest this word. We have adapted this revered learning process by introducing hints in various ways for advanced learning as well as competitive gaming.

\begin{itemize}
\item The \textbf{flashcards} may be the most general use of hints in learning on one side of which one writes a question and an answer overleaf. The learners can use it in either way to reinforce their learning outcomes by using the information supplied on one side of the card to recall that of on the other. When two people are using the flashcards at the same time, one usually uses hints to indirectly suggest the other to the answer to the question, and vice versa. 
\item The \textbf{spelling bee competition} involves another way that introduces hints to competitive gaming. During the contest, a word is usually pronounced first by the judge; then the contestant is expected to spell out the word accurately. However, when it becomes difficult, the contestant is usually allowed to be supplied with hints upon request that give some information about its spelling, such as the origin, meaning, or usage of the word for examples. 
\item The game of \textbf{hangman} involves a different hinting process as the gamer is expected to guess the spelling of the word within a limited number of trials. Unlike the above two, the verbal hint is optional, but the gamer is updated each time s/he makes a guess about the letters in the word, which provides evolving hints to narrow down the correct spelling. However, it is possible for the gamer to finish the game with complete \textit{lucky} guessing, or a strategic process by guessing the letters with a higher frequency of appearances in words. 
\end{itemize}

Proper design and use of hints can significantly influence the learning and practice of new knowledge. In general, hints are direct or indirect suggestions. One can conclude a few general guidelines when generating hints.
\begin{enumerate}
\item During the learning process, one can generate hints from the labeled outcomes through an indirect concept abstraction, 
\item Well-designed hints are preferred to be mutually exclusive and collectively exhaustive to suggest all possible outcomes without causing confusions to the learners, and 
\item The learners, after the learning process, can logically deduce the most likely answer by exhausting all possible hints in combination with the knowledge learned. 
\end{enumerate}
In this paper, we argue that one can transfer the above principals into the design of a hybrid neural network through the introduction of \textit{hints} as the auxiliary input, namely \textit{the indicators}. These indicators introduce a dynamic process of logical reasoning during training and testing, which is very different from the most of the apparent neural networks that directly correlate the input data and output labels. The logical reasoning process with the indicators, i.e. the hints, provide the hybrid neural network a self-checking process on the logic of the outcome, which improves the credibility of the network outcome during application with inevitable uncertainties. 

\begin{figure}[htbp]
\centering\includegraphics[width=1.0\linewidth]{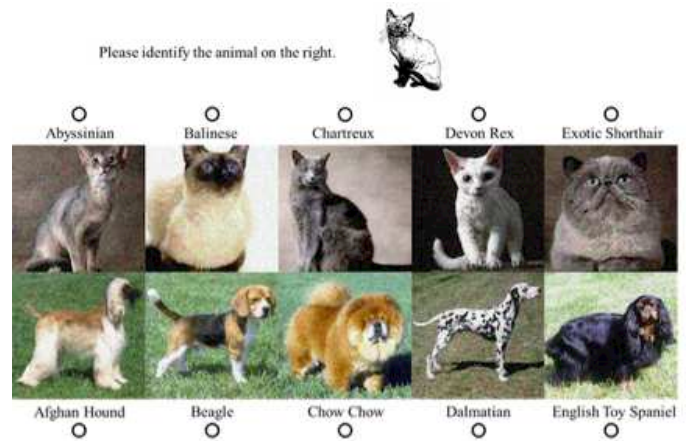}
\caption{A multi-choice question asking to recognize the animal on top by making a choice from the listed options.}
\label{FigCatOrDog}
\end{figure}

For example, in a classification problem where an animal lover is asked to recognize the breed of the animal from a selection of pictures labeled with five breeds of cats and dogs, as shown in Fig. \ref{FigCatOrDog}. This animal lover can choose to \textit{directly learn} how to recognize each of these animal breeds to determine what kind of cat or dog it is in this new picture. Alternatively, s/he can also choose to \textit{indirectly learn} under the hint of cats and dogs to aid the learning process within some logical reasoning, i.e. those breeds identified as cats shall not be confused with that of as dogs in the new knowledge learned. When presented with a new picture, this animal lover can choose to perform a direct recognition using the breeds of the ten animals learned. Alternatively, s/he can also apply the hints of cats and dogs while processing this new picture to find the most probable outcome that complies with results computed from all hints available. The latter shares more similarities with the human reasoning process, as one would usually start the explanation by saying that ``this is (probably) a Balinese cat (according to my learned knowledge of those ten animal breeds).'' In practice, one would also add to his/her explanation by saying that ``this is not a breed of dog.'' Through this simple self-checking mechanism, our animal lover can easily arrive at a confident conclusion about the animal breed in this picture. In practice, human reasoning usually processes the classification problem as a multiple-choice one based on the knowledge learned and a set of hints that indirectly corresponds to all possible choices, which are usually a finite, highly abstracted set of activities and daily life. This is the type of reasoning process that our proposed hybrid neural network intends to mimic.

For all problems dealing with the inevitable uncertainty of a new input data, there should be a new label of $y_{conflict}$ that raises questions when inconsistent results are computed. The human reasons in a similar way when we reach a contrary conclusion from a given input of vague clarity or confusing information. One would naturally question the answer rather than give an illogical one. Our research suggests an interactive interpretation and use of the label data as both input (i.e. hints) and output without compromising the learning outcome. The abstraction of the indicators can potentially play a more important role in the understanding of the hidden information in the input data. Unlike the direct use of input, our hybrid method enables one to provide suggestive data that helps the neural network to converge faster using less input. The generation process for the indicators can be a meaningful one, or randomized, depending on the applied design principles. When certain underlying patterns are observable in the labels, the introduction of the indicators as the auxiliary input will contribute as added information that helps the learning efficiency in a desirable way. However, when such underlying patterns are not clear, one can still use this hybrid structure as an exploratory method to experiment with data of a relatively small size, which is particularly useful when the cost of acquiring the data is relatively high.

\begin{figure*}[htbp]
\centering\includegraphics[width=0.8\linewidth]{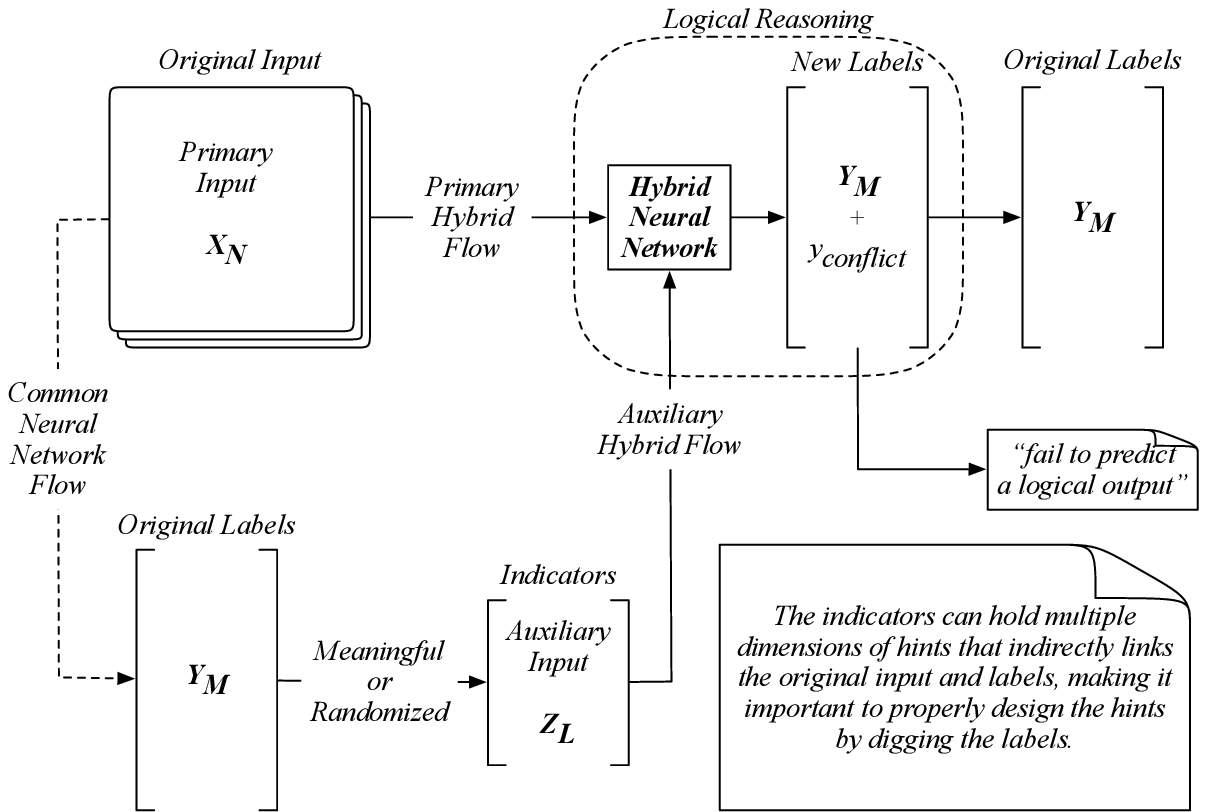}
\caption{A hybrid neural network with additional auxiliary input generated from the labels to improve the learning efficiency.}
\label{FigHybridNN_Learning}
\end{figure*}

The next section further explores the design variations of the indicators as the auxiliary input for the proposed hybrid neural network structure using the MNIST example. Section 3 compares the experiment results and discusses the design principles when generating these indicators from the labels. Section 4 further discusses an ongoing research about the application of this hybrid neural network in a logical learning setup for robotic grasping, namely the as\_DeepClaw. Final remarks and future work are enclosed in the last section, which end this paper.

%%%%%%%%%%%%%%%%%%%%%%%%%%%%%%%%%%%%%%%%%%%%%%%%%%%%
\section{The Design of a Hybrid Neural Network}
%%%%%%%%%%%%%%%%%%%%%%%%%%%%%%%%%%%%%%%%%%%%%%%%%%%%
The introduction of hints is the core differentiation behind our proposed hybrid neural network, which aims at a series of direct or indirect suggestions bridging the input data and the output labels with a reasonable decision. Since these indicators are suggestions abstracted from the originally labeled output, then a revisit to the information embedded in these labels becomes essential. In general, the labeled outcome usually correspond to a comprehensive and complex question that is hard to answer and therefore cannot be easily modeled using the existing knowledge. Although the final reply to the question may be framed as a simple \textit{Yes} or \textit{No}, one could usually break down the question to further details for more specific inquiry. For example, the question asked in Fig. \ref{FigCatOrDog} can be divided into 1) \textit{what is the animal in general} and 2) \textit{what is the breed of this animal specifically}. Such operation may have different logical complexities, just like when people are trying to find solutions to the same problem, some may follow the existing procedures that are straightforward but difficult in practice, while others might \textit{``think out of the box''} and identify particular patterns from the questions and possible answers to formulate a logical process that leads to a much-simplified solution. Our hybrid neural network is trying to mimic such deductive human intelligence using these hints. 

%%%%%%%%%%%%%%%%%%%%%%%%%%%%%%%%%%%%%%%%%%%%%%%%%%%%
	\subsection{The Auxiliary Input and New Labels}
%%%%%%%%%%%%%%%%%%%%%%%%%%%%%%%%%%%%%%%%%%%%%%%%%%%%
Before our training starts, we are presented with a set of input data with labeled output as the \textit{prior information}. In our hybrid neural network shown in Fig. \ref{FigHybridNN_Learning}, the original input is still used as the primary input $\boldsymbol{X_{N}}=\left \{ x_1, x_2, ..., x_N \right \}$. In addition, we generate a set of auxiliary input to the network, namely the indicators, by categorizing the original labels $\boldsymbol{Y_{M}}=\left \{ y_1, y_2, ..., y_M \right \}$. Since the understanding of these labels is not directly linked to the input, we can exploit such \textit{indirect} knowledge that is not presented in the prior information to categorize these output labels, which can be a meaningful process or a randomized one. The resultant indirect suggestions become a set of indicators $\boldsymbol{Z_{L}}=\left \{ z_1, z_2, ..., z_L \right \}$ which usually has a smaller dimension than the original labels ($L < M$) as a sign of concept abstraction. For a logical modeling, any original input $x_i$ shall be led to its original output $y_i$ through the involvement of a particular indicator $z_i$ that indirectly suggest this correct computation. When computing this original input $x_i$ with all other indicators, a new label $y_{conflict}$ becomes necessary to differentiate this illogical outcomes from the logical ones. Therefore, we shall have a set of new labeled output $\boldsymbol{{Y}_{M+1}}=\left \{ y_1, y_2, ..., y_M, y_{conflict} \right \}$ when training our hybrid neural network. One can further exploit the concept of hints by designing multi-dimensional or direct indicators for alternative logics. One can also design a new set of restructured labels from the original input data for more advanced logical learning, which will be introduced in the later sections of this paper.

%%%%%%%%%%%%%%%%%%%%%%%%%%%%%%%%%%%%%%%%%%%%%%%%%%%%
	\subsection{New Architecture for Logical Learning}
%%%%%%%%%%%%%%%%%%%%%%%%%%%%%%%%%%%%%%%%%%%%%%%%%%%%
During the training process shown in Fig. \ref{FigHybridNN_Learning}, our resultant neural network takes a primary input $x_i$ and an auxiliary one $z_i$ to model a new set of labeled output $y_i$. This hybrid neural network is structurally different by adding an additional process that establishes a certain logical relationship in the original data. The advantage of the hybrid neural network is the full exploitation of our existing understandings of the problem through the design of meaningful indicators and the establishment of logical reasoning. Later we will also show that the training process can be further exploited by designing randomized indicators to suggest unknown relationships and reduce the data and logic uncertainty.

\begin{figure*}[htbp]
\centering\includegraphics[width=0.8\linewidth]{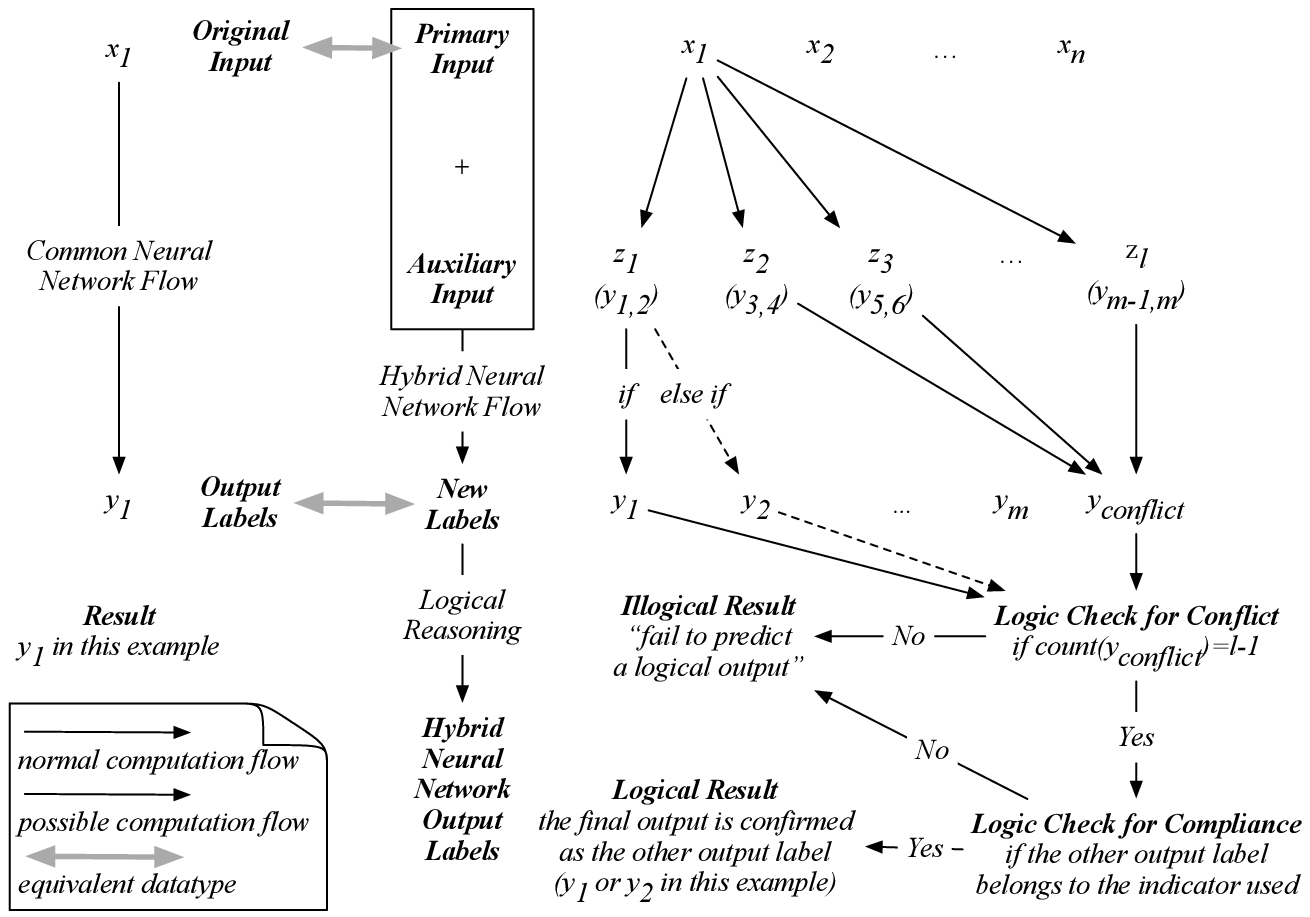}
\caption{A hybrid neural network with additional auxiliary input generated from the labels to improve the learning efficiency.}
\label{FigHybNN_Testing}
\end{figure*}

Since our auxiliary input is an artificial one, we are only provided with the same input data as in the apparent neural network to proceed with our prediction. For example, as shown in Fig. \ref{FigHybNN_Testing}, when a new input $x_1$ is presented, our hybrid neural network will exhaust all possible indicators in $\boldsymbol{Z_L}$. To logically determine a reasonable prediction, we can start by counting whether the number of $y_{conflict}$ predicted equals to $L-1$, which means that only one non-conflicting label is predicted after the exhaustion of all possible indicators. If so, the next logical check is to determine if the one-and-only non-conflicting label complies with the corresponding indicator used to predict this non-conflicting label. If the one-and-only non-conflicting label passes both logical reasoning tests, then a logical result is obtained by our hybrid neural network. Otherwise, an illogical result will be reached with a statement such as ``fail to predict a logical output.'' This is fundamentally different from the common apparent models where an output label will always be computed even at a low confidence without cross-checking the logic behind. 

%%%%%%%%%%%%%%%%%%%%%%%%%%%%%%%%%%%%%%%%%%%%%%%%%%%%
\subsection{The Adapted MNIST Data}
\begin{figure*}[htbp]
\centering\includegraphics[width=0.6\linewidth]{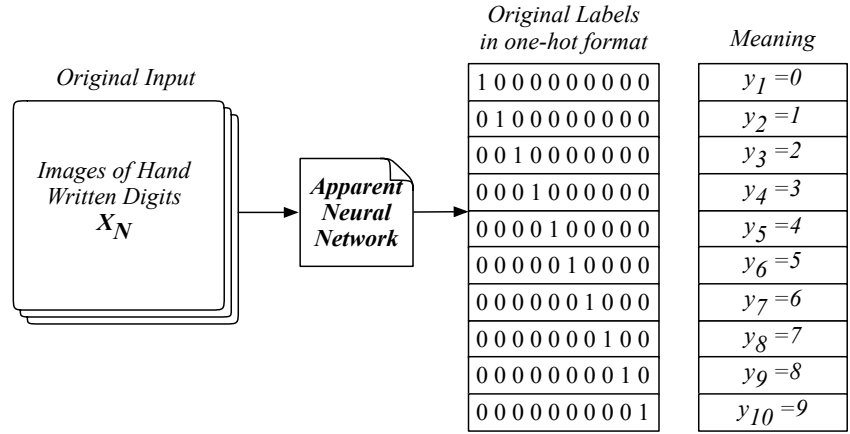}
\caption{The MNIST data and the apparent neural networks widely used in machine learning.}
\label{FigOriginalMNIST}
\end{figure*}

%%%%%%%%%%%%%%%%%%%%%%%%%%%%%%%%%%%%%%%%%%%%%%%%%%%%
The MNIST data is an extensive image collection of hand-written digits labeled from zero to nine, consisting of a training set of 60,000 examples and a test set of 10,000 examples  \citep{LeCun1998Gradient-basedRecognition}. It is widely adopted as a benchmarking dataset for machine learning research and practice. Researchers usually adopt the \textit{apparent neural networks} to use the images as the input data $\boldsymbol{X_{N \times 28 \times 28 \times 1}}$ and the digits as the output labels $\boldsymbol{Y_{10}}$ as shown in Fig. \ref{FigOriginalMNIST}. In practice, the model outcomes are effectively labeled in a one-hot format which corresponds to the ten different digits. These ten digits are treated as ten independent categories disregarding their mathematical meanings and relationships behind. This \textit{direct learning} method is comparable to the way we approach the dictionary for the meaning of a new word. However, it hardly reflects the logical and interactive way we naturally adopt to use words in a language, as reflected in the examples at the beginning of this paper, which requires the help from \textit{logical learning}.

\begin{figure*}[htbp]
\centering\includegraphics[width=0.6\linewidth]{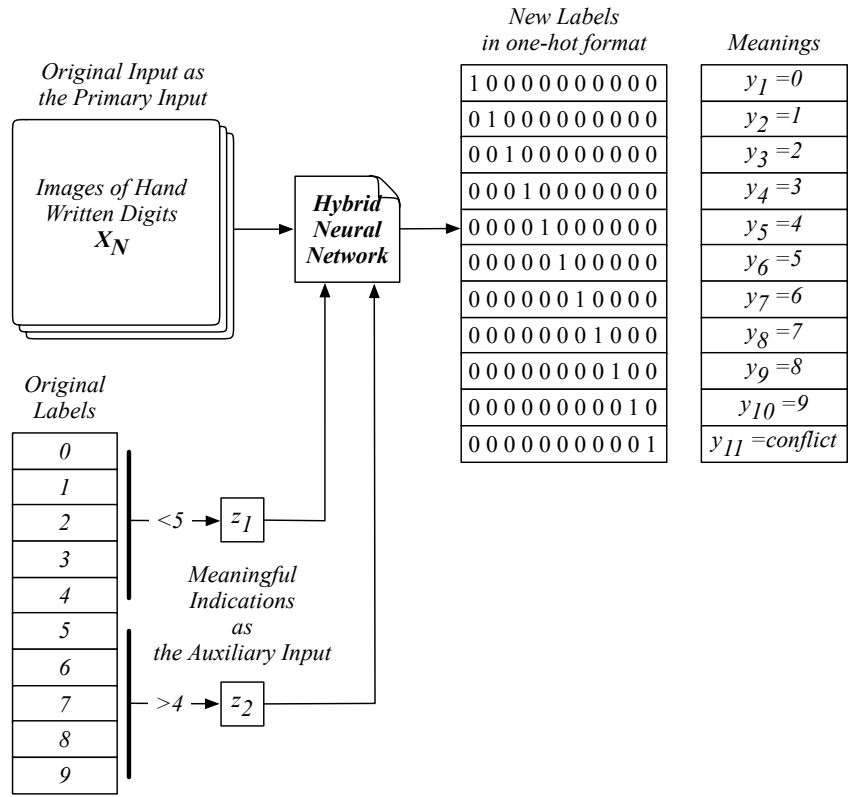}
\caption{A hybrid neural network for training with an adapted MNIST data with two indicators, suggesting original labels smaller than five or not, as the auxiliary input.}
\label{FigHybridMNIST_Learning}
\end{figure*}

The example in Fig. \ref{FigHybridMNIST_Learning} shows the training setup with a set of two indicators, with $z_1$ suggesting the label being smaller than five and $z_2$ suggesting otherwise. The training of this hybrid neural network involves the combination of the primary input and all possible indicators. For example, with an image of digit zero as the input, we generate the label when this image and indicator $z_1$ are used, which computes the correct label of $y_1=0$. In the meanwhile, we also generate another label when this image and the other indicator $z_2$ are used, which lead to the conflicting output label of $y_{conflict}$. This process essentially tells the hybrid neural network to learn both sides of the knowledge, including the direct and indirect ones. The introduction of this new label $y_{conflict}$ enables our hybrid neural network to capture the logics of this new knowledge during the training process, which will be used for the logical reasoning during the testing and application of this hybrid neural network. 

\begin{figure*}[htbp]
\centering\includegraphics[width=0.6\linewidth]{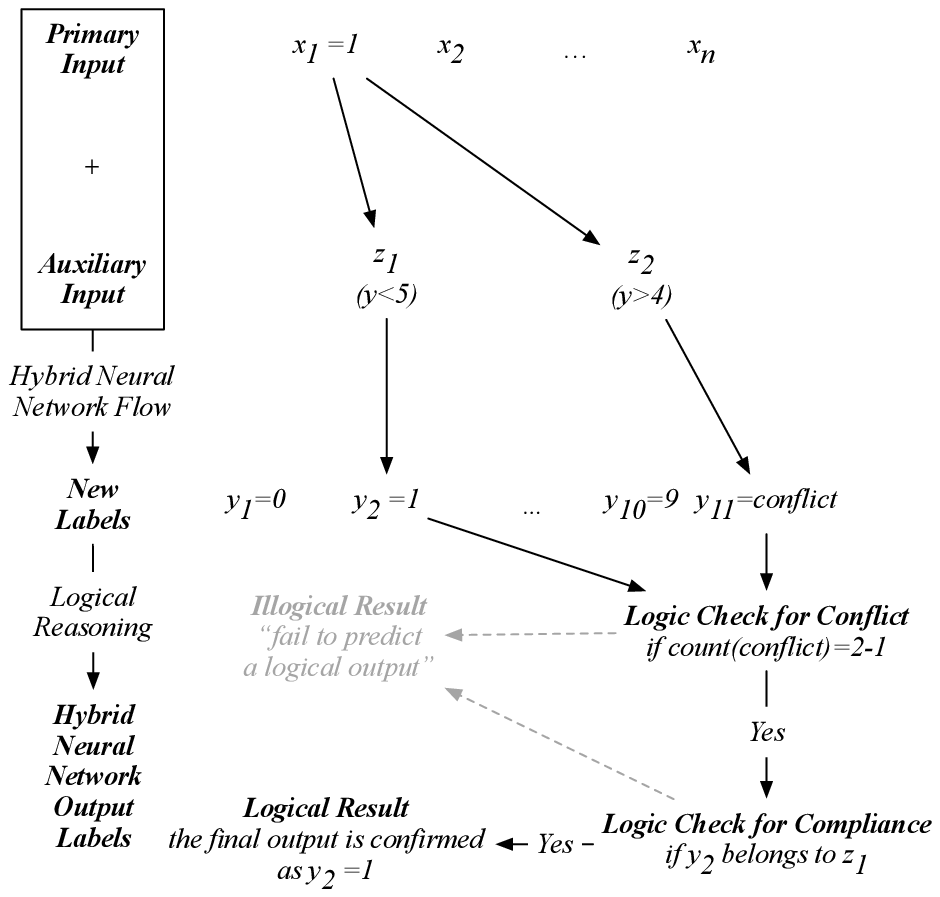}
\caption{The example showing the testing of the hybrid neural network for a logical output.}
\label{FigHybMNIST_Testing}
\end{figure*}

The testing of this hybrid neural network is also very different from the apparent ones, which involves an exhaustive computation between a given primary input and all possible indicators, and a logical reasoning process trying to make sense out of the predicted results. The example in Fig. \ref{FigHybMNIST_Testing} demonstrates the case when a image $x_1$ of digit 1 is used for testing. The hybrid neural network flow computes $L=2$ label outputs that correspond to all combinations of $x_1$ and $z_1$, and $x_1$ and $z_2$ as the input pair. Then, in the logical reasoning flow, we firstly check if the total number of $y_{conflict}$ predicted equals to $L-1$, meaning that only one non-conflicting label is predicted. Then, we further check if this non-conflicting label complies with the corresponding indicator used for its computation. A logical result can only be obtained when a non-conflicting label is in compliance with the indicator used for its computation. If the set of predicted labels fail to pass any of these two logical checks, we can still arrive at an output label by selecting those with lower computation confidence, just like the method used in the apparent neural networks. However, this will only result in illogical results, which the apparent neural network can not detect. An extreme example is when one supplies an entirely irrelevant image to a neural network, like a picture of a cat instead of hand-written digit. The apparent neural network will process it with a labeled output, possibly with a very low confidence, asking for manual checking. This corresponds to a statement such as ``the most probable recognition of this new image is a digit ...''. However, our proposed hybrid neural network can identify all possible labels computed to collectively suggest an illogical output, which corresponds to a statement such as ``no logical recognition can be computed,'' suggesting something is wrong with this image or the model currently in use for this task based on the knowledge learned. 

For demonstration, the original MNIST data has been adapted in various ways in the following section by introducing different sets of indicators to restructure the neural network. In principal, we can design a few sets of the indicators to indirectly suggest the ten digits according to 1) the meaning of the indicators, 2) the total number of indicators, and 3) the number of labels suggested by each indicator. 

We first design four sets of indicators numbered as 11 to 14 in Table \ref{TabHybNormal} with mathematical meanings. The ten digits are divided into two categories, including 1) smaller than 5 or not (equally divided), 2) even number or not (equally divided), 3) prime number or not (unequally divided), and 4) zero or not (unequally divided). We also randomly generate four sets of indicators numbered as 21 to 24 in Table \ref{TabHybNormal} with different total counts $L$ and size distributions. All sets of indicators in Table \ref{TabHybNormal} provide indirect suggestions to all ten digit labels except for cases 14 and 24. Case 14 contains a direct suggestion of an indicator zero to label zero, and case 24 contains direct suggestions of the ten indicators to the ten digit labels respectively. Both designs of the indicators violate the general design guidelines of hints as defined earlier. More specifically, case 14 represents a partial violation whereas case 24 accounts for a total breach. However, whether such violation is a bad thing remains an open question, which will be discussed later.

\begin{table*}[htbp]
\centering
\begin{tabular}{ c c c c }
\hline 
\multirow{2}{*}{\textbf{\#}} & \textbf{Meanings of} & \multicolumn{2}{c}{\textbf{Indicator Characteristics}}\tabularnewline
\cline{3-4} 
 & \textbf{the Indicators} & \textbf{Total Count $L$ } & \textbf{Size Distribution}\tabularnewline
\hline 
11 & Smaller than 5 or not	& 2     & Equal / 5-5    \tabularnewline
12 & Even number or not     & 2     & Equal / 5-5    \tabularnewline
13 & Prime number or not	& 2     & Unequal / 4-6    \tabularnewline
14 & Zero or not			& 2     & Unequal / 1-9    \tabularnewline
21 & None					& 2     & Equal / 5-5    \tabularnewline
22 & None					& 2     & Unequal / 3-7    \tabularnewline
23 & None					& 5     & Equal / 2$\times$5    \tabularnewline
24 & None					& 10	& Equal / 1$\times$10    \tabularnewline
\hline 
\end{tabular}
\caption{Normal design examples of the auxiliary inputs for the MNIST data.}
\label{TabHybNormal}
\end{table*}

Since the indicators only provide indirect suggestions, we can also design new prediction models as shown in Table \ref{TabHybSpecial} by interchanging the labels and the indicators used in the hybrid neural networks. One special design of the indicators is a multi-dimensional indirect suggestion. For example, each digit can be simultaneously identified as smaller than five and an even number. This corresponds to case 31 in Table \ref{TabHybSpecial} where a set of two-dimensional indicators carries the meanings of both cases 11 and 12. Another special design is through the exchange of the indicators and labels to formulate a revered prediction, as shown in case 32 in Table \ref{TabHybSpecial}. A further special design is to hide the exact information of the ten digit by using indirect suggestions in both indicators and labels, such as the case 33 in Table \ref{TabHybSpecial}. Note that in this case, since we are losing the exact information of the specific digit on the image, we can only do the first layer of logic check for conflicts but cannot proceed to the second layer of logic check for compliance.

\begin{table*}[htbp]
\centering
\begin{tabular}{ c c c c }
\hline 
\# & \textbf{Special Indicator Design} & \textbf{auxiliary Input} & \textbf{Output Labels}\tabularnewline
\hline 
31 & Multi-dimensional Indication	& 2D (Case 11+Case12)	& ten digits\tabularnewline
32 & Reversed Prediction			& ten digits			& Case 11\tabularnewline
33 & Irrelevant Prediction			& Case 12				& 11\tabularnewline
\hline 
\end{tabular}
\caption{Special designs of the hybrid neural network using the adapted MNSIT data for model prediction.}
\label{TabHybSpecial}
\end{table*}

%%%%%%%%%%%%%%%%%%%%%%%%%%%%%%%%%%%%%%%%%%%%%%%%%%%%
\section{Experiment Results and Discussions}
%%%%%%%%%%%%%%%%%%%%%%%%%%%%%%%%%%%%%%%%%%%%%%%%%%%%
We benchmark our models with the MNIST data using a Convolutional Neural Network that contains three convolutional layers ($5\times5\times1\times4$ with stride 1, $4\times4\times4\times8$ with stride 2, and $4\times4\times8\times12$ with stride 2) and two fully connected layers ($588\times200$ and $200\times10$). The test accuracy of the benchmark model maximizes at 98.91\% after 10,000 training steps with training batch equals to 100. One can further improve the accuracy by referring to \citep{Gorner2016LearnPh.D.} using more advanced techniques such as more layers, batch normalization, dropout and ReLu layers.

All experiments with indicators listed in Table \ref{TabHybNormal} are conducted \citep{Wan2017Github}, and the testing results are reported in Table \ref{TabHydResultNormal}. The \textit{Logical Results} column refers to the percentage of the 10000 MINIST testing examples that passing the two rounds of logical checks. This self-checking metric leads to two prediction accuracies of interest. One is the prediction accuracy of the \textit{Logical Results}, while the other is that of the \textit{Overall Results}. Those \textit{Illogical Results} are marked as failed recognition, adding to the overall results. Note that the Prediction Accuracy of the Overall Results can be considered as equivalent to testing accuracy in the apparent neural networks.

\begin{table}[htbp]
\centering
\begin{tabular}{ c c c c }
\hline 
\multirow{2}{*}{\textbf{\#}} & \textbf{Logical} & \multicolumn{2}{c}{\textbf{Prediction Accuracy}}\tabularnewline
\cline{3-4} 
 & \textbf{Results} & \textbf{Logical} & \textbf{Overall}\tabularnewline
\hline 
11 & 99.48\% & 99.24\% & 98.72\%\tabularnewline
12 & 99.49\% & 99.21\% & 98.70\%\tabularnewline
13 & 99.57\% & 99.13\% & 98.70\%\tabularnewline
14 & 99.90\% & 99.00\% & 98.90\%\tabularnewline
21 & 99.36\% & 99.17\% & 98.54\%\tabularnewline
22 & 99.40\% & 99.32\% & 98.72\%\tabularnewline
23 & 98.79\% & 99.39\% & 98.19\%\tabularnewline
24 & 98.53\% & 99.41\% & 97.95\%\tabularnewline
\hline 
\end{tabular}
\caption{Model prediction results of the normal hybrid neural networks.}
\label{TabHydResultNormal}
\end{table}

%%%%%%%%%%%%%%%%%%%%%%%%%%%%%%%%%%%%%%%%%%%%%%%%%%%%
\subsection{Any indicator is a good indicator for logical reasoning}
%%%%%%%%%%%%%%%%%%%%%%%%%%%%%%%%%%%%%%%%%%%%%%%%%%%%
As shown in Table \ref{TabHydResultNormal}, all logical results are above 99\% prediction accuracy, which is higher than the apparent neural network model's benchmarking prediction accuracy at 98.91\%. Irrespective of the indicator designs, our results suggested reassured confidence when a logical answer is concluded. This is a self-checking mechanism of the hybrid neural network, which is not available in the apparent neural networks.

Apparent neural networks only adopt the direct information presented in the data but ignores the logics behind the data, which is usually a reasoning process of human thinking instead of memorizing past events using brutal computation. These new logics help to deal with the uncertainties in future events. Our hybrid neural network employs a logical learning process through the introduction of the indicators for suggestive information. The robustness of the results shown in Table \ref{TabHydResultNormal} demonstrates the effectiveness of the proposed logical learning through the hybrid neural network. These indicators are flexible in design to provide direct (cases 11-13 and 21-23) or indirect (cases 14 and 24) suggestions for the labeled outcome. Moreover, the designs and understandings of these indicators are the prior knowledge that the human operators had previously acquired from the given dataset for artificial learning.

%%%%%%%%%%%%%%%%%%%%%%%%%%%%%%%%%%%%%%%%%%%%%%%%%%%%
\subsection{Logical complexity positively relates to the confidence of a logical answer}
%%%%%%%%%%%%%%%%%%%%%%%%%%%%%%%%%%%%%%%%%%%%%%%%%%%%
As shown in Table \ref{TabHybNormal}, three levels of the concept are used to reflect different logical complexities when designing the indicators. Those generated from random partially represents a certain unknown relationship with the highest complexity, whereas the meaningful ones are more straightforward for human understanding. The count of indicators is also an important aspect to reflect the complexity of the logical reasoning in our models. For example, every added indicator would boost the required training by one fold as we traverse all the indicators. Furthermore, the distribution of each indicator and the labels it suggests also present a statistical influence that correlates to the distribution of the input data. 

\begin{figure}[htbp]
\centering\includegraphics[width=1.0\linewidth]{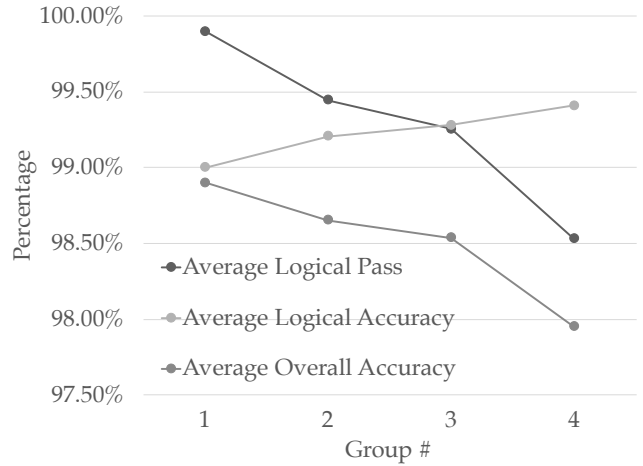}
\caption{The average results of the four case groups of the normal indicators used.}
\label{FigNormalGroup}
\end{figure}

In general, we can divide these eight experiments into four groups according to their logic complexities. As shown in Fig. \ref{FigNormalGroup}, a trend of growing logical accuracy is observed as the logic complexity increases. The group with the least logical complexity only contains case 14, which shares the most similarity to the original MNIST dataset with the lowest prediction accuracy of 99.00\% for the logical results. This is effectively a direct indication for images of zero and a reduced MNIST prediction without the zeros. Such simplicity on logic leads to the highest percentage of passing the logical checks at 99.90\%. 

The second group contains cases 11, 12, and 21 with slightly more complex logical relationships, which are either meaningful or randomized indicators and all have a set of two indicators suggesting an equal number of labels. An average logical accuracy of 99.21\% is reached with the percentage of passing the logical checks reduced slightly to 99.44\%. 

The third group consisting of cases 13, 22, and 23 presents more complexity in all three aspects of indicative meaning, indicator count, and size distribution. For example, both cases 13 and 23 have unequal size distributions comparing to Group 2. Moreover, the total count of indicators increases to five even though the size distribution stays equal. This group achieves an increased average logical accuracy of 99.28\% with a  99.25\% passing rate of the logical checks.

The fourth group only contains case 24, which is the most complex logic with ten indicators of direct suggestions for training. It produces the highest logical accuracy of 99.41\%. The side effect of such restrict logic is that the passing rate of the logical checks only 98.53\%, the lowest in our experiments. 

%%%%%%%%%%%%%%%%%%%%%%%%%%%%%%%%%%%%%%%%%%%%%%%%%%%%
\subsection{Logical result is at the cost of overall accuracy}
%%%%%%%%%%%%%%%%%%%%%%%%%%%%%%%%%%%%%%%%%%%%%%%%%%%%
While the logical result offers us with a more confident answer when dealing with the unknown uncertainties of a new question, i.e. a new input data, the prediction accuracy of the overall result provides us with a way to compare the hybrid neural networks with the apparent ones horizontally. We also plotted the trend of overall prediction accuracies of the aforementioned four experiment groups in Fig. \ref{FigNormalGroup}. A decreasing trend can be observed as the accuracy of the logical results increases. In fact, the percentage of logical pass shows a stronger influence on the overall prediction accuracy in Fig. \ref{FigNormalGroup}. This is caused by the fact that the decrease in the logical pass is about three times faster than the increase in logical accuracy. The following equation can be used to reflect such relationship.
\begin{equation}
\label{eq:emc}
Overall Accuracy = Logical Accuracy \times Logical Pass
\end{equation}

%%%%%%%%%%%%%%%%%%%%%%%%%%%%%%%%%%%%%%%%%%%%%%%%%%%%
\subsection{Meaningful indicators are not that good, and direct suggestions are not that bad}
%%%%%%%%%%%%%%%%%%%%%%%%%%%%%%%%%%%%%%%%%%%%%%%%%%%%
Another observation is about the design of the indicators about their meanings and suggestive strength. Comparing cases 11, 12, and 21, we can see that the meanings of the indicators only present a slight increase in the prediction accuracy of the logical results from 99.17\% to 99.21-99.24\%. One probable explanation is that the computers do not perceive such mathematical relationships much differently from the randomized ones, which is very different from the humans. This partially reflects the underlying difference between the humans and the machines, where such slight increase in the understanding makes a significant difference in the autonomy of minds. 

\begin{figure}[htbp]
\centering\includegraphics[width=1.0\linewidth]{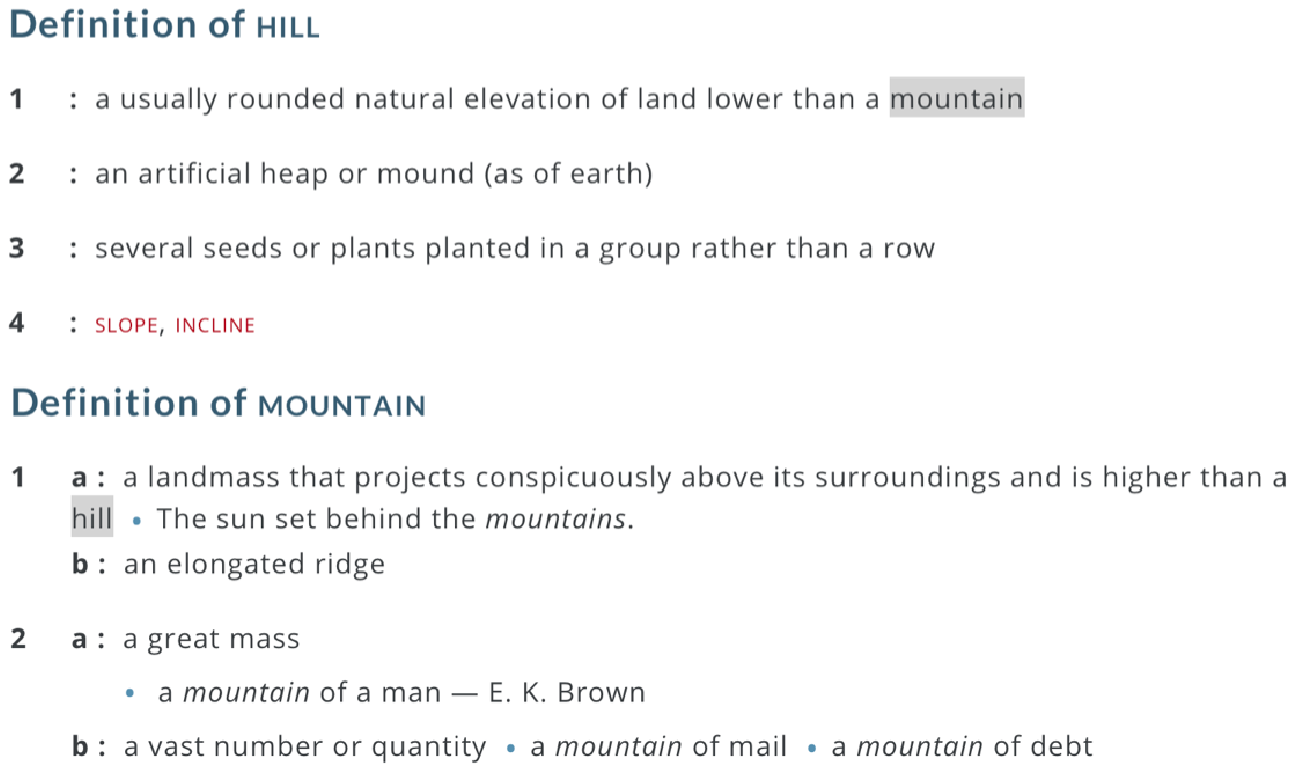}
\caption{Example of a circular definitions of hill and mountain from the Merriam-Webster online dictionary.}
\label{FigCircularDef}
\end{figure}

While designing these indicators, we intentionally tried to mimic the human use of hints in an indirect way as introduced at the beginning of this paper. Among the eight normal experiments, only cases 14 and 24 involves direct 1-to-1 suggestions between the indicators and the labels. These two cases outperformed the rest ones in logical accuracy. All eight experiments present an average logical prediction accuracy at 99.23\% (0.0014 standard deviations) and an average overall prediction accuracy of 98.55\% (0.0032 standard deviations) that are uniform in results. This reflects the fact that it is challenging to draw a clear line between direct and indirect suggestion. We can refer to the example of the circular definitions of the dictionaries as demonstrated in Fig. \ref{FigCircularDef}. It is obviously not appropriate to use the same word to explain itself. However, when other words are used for an explanation, it could suggestion both direct or indirect meanings of the word, as long as it helps the language learner to relate the word to a comprehensive meaning for understanding. 

%%%%%%%%%%%%%%%%%%%%%%%%%%%%%%%%%%%%%%%%%%%%%%%%%%%%
\subsection{Special Design and Use of the Indicators}
%%%%%%%%%%%%%%%%%%%%%%%%%%%%%%%%%%%%%%%%%%%%%%%%%%%%
We also list a few special cases in Table \ref{TabHybSpecial} to demonstrate the versatile design and use of the indicators in a hybrid neural network, with interesting results reported in Table \ref{TabHydResultSpecial}. 

Case 31 is a two-dimensional design of the indicators which combines case 11 and case 12. The result shows an increase in the prediction accuracy of the logical result of 99.33\%, but a decrease in the passing rate of the logical check to 98.97\%. In principle, it is possible to design such multi-dimensional indicators. However, the major difference is more operational instead of numerical. Multi-dimensional indicators require the each input to be supplied with multiple indicators instead of one as shown in Fig. \ref{FigHybridNN_Learning}. This design is equivalent to four one-dimensional indicators.

Case 32 is a reversed prediction by exchanging the labels and indicators in case 11. The major interest of this case is the application of the hybrid neural network when the number of the indicators are larger than the number of labels. In this case, we use ten meaningful indicators to suggest two possible labels given an input data. The passing rate of logical check is found to be the smallest among all experiments, suggesting the highest logical complexity and resulting in the lowest overall prediction accuracy. However, we did not suffer any significant decrease in logical accuracy, which remains at a relatively high level of 99.36\%. During operation, this case will require the same amount of computations as case 24 to traverse all indicators. By applying the conclusion of \textit{``meaningful indicators are not that good, and direct suggestions are not that bad''} obtained in the last section, one can effectively introduce a random set of indicators whose size is bigger than that of the output labels to reach a reasonably high prediction accuracy of the logical results. This is especially useful when the cost of data acquisition is high. One example of such use scenario is the autonomous robotic grasping task, which will be further expanded with more details in the next section. 

The last case 33 completely buries the explicit information of the ten numerical digits in the data. The overall accuracy is the highest among all normal and special hybrid neural networks at 98.91\%. In fact, this result coincides with the apparent neural network's testing accuracy. The result for case 33 suggests that the use of hints may be the most suitable for indirect suggestions to effectively address unknown uncertainties and find new knowledge that was not previously expressed in the original data. One drawback of case 33 is that the second logic check cannot be performed as we cannot decide if a number smaller than five is an even number or not. Further research is required to dive further into this kind of problem. 

\begin{table}[htbp]
\centering
\begin{tabular}{ c c c c }
\hline 
\multirow{2}{*}{\textbf{\#}} & \textbf{Logical} & \multicolumn{2}{c}{\textbf{Prediction Accuracy}}\tabularnewline
\cline{3-4} 
 & \textbf{Results} & \textbf{Logical} & \textbf{Overall}\tabularnewline
\hline 
31 & 98.97\% & 99.33\% & 98.31\%\tabularnewline
32 & 97.33\% & 99.36\% & 96.71\%\tabularnewline 
33 & 99.58\% & 99.33\% & 98.91\%\tabularnewline
\hline 
\end{tabular}
\caption{Model prediction results of the special hybrid neural networks.}
\label{TabHydResultSpecial}
\end{table}

%%%%%%%%%%%%%%%%%%%%%%%%%%%%%%%%%%%%%%%%%%%%%%%%%%%%
\section{Logical Learning for the as\_DeepClaw}
%%%%%%%%%%%%%%%%%%%%%%%%%%%%%%%%%%%%%%%%%%%%%%%%%%%%
In this section, we dive further into the application of the logical learning using the hybrid neural network for autonomous robotic grasping. The robotic platform is set up as an arcade claw machine, i.e. as\_DeepClaw, with more details described on a GitHub repository \citep{Wan2017As_DeepClaw:Network}. It involves a general purpose six-axis robot arm performing grasping tasks with stuffed toys as the target objects and an RGB-D camera as the sensory input. Different from previous research using deep learning methods for autonomous robotic grasping \citep{Levine2016LearningCollection, Pinto2015SupersizingHours, Lenz2013DeepGrasps}, we intend to apply the logical learning using the hybrid neural network for training with the as\_DeepClaw platform. The arcade claw machine can be considered as a structurally simplified general purpose grasping setup implemented with a complete robotic system. The design of the particular grasping process mimics the actual arcade claw machine game as much as possible.

\begin{figure}[htbp]
\centering\includegraphics[width=0.7\linewidth]{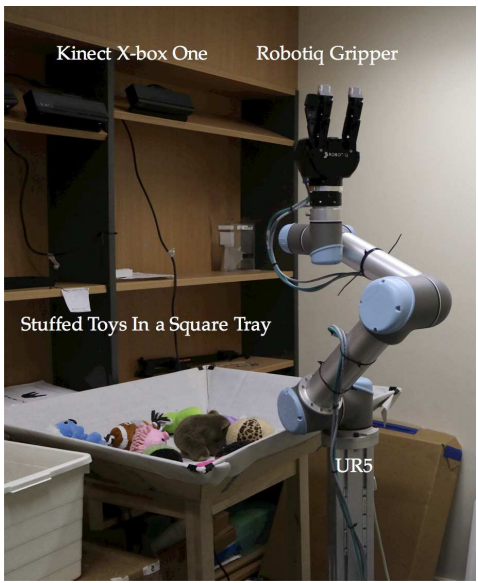}
\caption{The setup of as\_DeepClaw that mimics an arcade claw machine using an UR5 arm, a Robotiq 3 Finger Gripper, two Kinect for Xbox One cameras, and a PC that with NVidia Titan X 12G GPU to pick up stuffed toys from a tray.}
\label{FigAsDeepClaw}
\end{figure}

The training data is collected through \textit{blind grasping} by randomly choose a probable coordinate $(x, y)$ on the object plane a few centimeters above the bottom of the tray. The gripper is maintained at a vertical pose at all times but is allowed to rotate about the vertical axis with $\theta$. The cameras take pictures of the tray as the major sensory input. An apparent neural network for this task would be similar to those described in \citep{Levine2016LearningCollection, Pinto2015SupersizingHours,Lenz2013DeepGrasps}. The original input will be the images taken from the tray, and the output labels will be the grasping success or failure.

We performed some adaptations on the input and output data to experiment this robotic setup using logical learning with a hybrid neural network. We adapted the primary input data as the cropped images based on the average size of the stuffed toys and the coordinates of the gripper $(x, y)$ before the blind grasping. This is a way to simplify our training difficulty so that the coordinate information is indirectly buried inside each cropped image. This leaves us with only the rotational angle $\theta$ as the auxiliary input, which we discrete into steps of ten degrees per indicator. Besides, we also introduced an indicator to suggest grasping uncertainty for a relaxation of the logical checks for the input data. During actual experiment, many uncertainties such as gripper finger friction, the passive operation of the Robotiq gripper, or the fluffy surface of the stuffed toys, etc. may cause the mislabeling of the grasping outcome, which is adjusted through this indicator. 

As for the adapted output labels, we performed some extra work by extracting object information from the cropped input images and add it to the original output labels. Each kind of the stuffed toy corresponds to three output labels, including one for successful grasping, one for failed grasping, and one for untested grasping. A conflict label is still added to the training output for logical checking. Overall, this logical learning structure is more complex in logical reasoning, which reflects the difficult nature of the autonomous grasping tasks. This is an ongoing research and will be separately addressed in another paper with more details. 

%%%%%%%%%%%%%%%%%%%%%%%%%%%%%%%%%%%%%%%%%%%%%%%%%%%%
\section{Final Remarks and Future Work}
%%%%%%%%%%%%%%%%%%%%%%%%%%%%%%%%%%%%%%%%%%%%%%%%%%%%
In this paper, we propose the concept of logical learning through a hybrid neural network with an additional auxiliary input, namely indicators, generated from the original labels, or even the original primary input data. Given the same data, the logical learning can always provide results with a higher logical accuracy that is supported by a logical reasoning process. We comprehensively introduced the concept of indicators for logical learning with the hybrid neural network on its design and use during model training as well as prediction. We also demonstrated the robustness of the proposed logical learning in a series of normal and special indicators. A few guidelines are summarized below to help assist the design and use of the indicators in a hybrid neural network. 
\begin{itemize}
\item Any indicator is a good indicator for logical reasoning;
\item Logical complexity positively relates to the confidence of an answer;
\item Logical result is at the cost of overall accuracy;
\item Meaningful indicators are not that good, and direct suggestions are not that bad;
\item The design of the indicators is not limited by our understanding of the data.
\end{itemize}

This powerful tool provides us with a way to reflect the logical reasoning process while trying to comprehend more advanced concepts. It enables us to model the \textit{unknown unknowns} structurally without the loss of confidence when a new and uncertain input is supplied. This process can be a meaningful one through the design of the indicators when we have established some prior understanding of the data, or a randomized one when we only care about a possible logic for a reasonable answer instead of the \textit{why} behind. 

The advancement of computing capabilities enables us to do brutal computation using neural networks for a most probable answer without caring much in the logic behind the answer. The average percentage of passing logical checks is at a relatively high level of 99.32\%, leaving only a small fraction of data marked as illogical. However, this is mainly due to the high quality and simplicity of the MNIST data, which might not be the case for other machine learning tasks. It is particularly challenging when the cost of getting training data is expensive, especially when data collection requires physical interactions with the external environment, such as the robotics. When only a limited amount of data is available, it is essential to utilize all aspects of the data, including the logical reasoning, physical meaning, as well as environmental variables, etc., for a potential solution with the most probable confidence. 

While the scope of this paper is to introduce the concept of logical learning with a hybrid neural network through auxiliary input as the indicators, future work requires a systematic research into the comprehensive and logical design of the indicators and hybrid neural network structures. The ongoing research of the as\_DeepClaw is an immediate direction to test further the application of the logical learning, which will be reported separately in an upcoming paper.

% \lipsum[1-3]

% \begin{itemize}
% \item Bullet point one
% \item Bullet point two
% \end{itemize}

% \begin{enumerate}
% \item Numbered list item one
% \item Numbered list item two
% \end{enumerate}

% \subsection{Subsection One}

% \lipsum[2-1]

% \begin{table}[h]
% \centering
% \begin{tabular}{l l l}
% \hline
% \textbf{Treatments} & \textbf{Response 1} & \textbf{Response 2}\\
% \hline
% Treatment 1 & 0.0003262 & 0.562 \\
% Treatment 2 & 0.0015681 & 0.910 \\
% Treatment 3 & 0.0009271 & 0.296 \\
% \hline
% \end{tabular}
% \caption{Table caption}
% \end{table}

% \subsection{Subsection Two}

% \lipsum[3-1]

% \begin{figure}[h]
% \centering\includegraphics[width=0.4\linewidth]{placeholder}
% \caption{Figure caption}
% \end{figure}

% \lipsum[4-1]

% \begin{equation}
% \label{eq:emc}
% e = mc^2
% \end{equation}

\section*{Acknowledgment}
The authors would like to acknowledge their daughter, SONG Roumu, for her growing capabilities of learning that inspired this research. 

The authors would like to also acknowledge the following students visiting the Sustainable and Intelligent Robotics Group led by the corresponding author at the Monash University, who had worked on setting up the as\_DeepClaw robotic system, including XIA Tian, HE Xiaoyi, and DENG Yuntian.

\section*{References}

% % Please connect to your Mendeley account and use the appropriate Mendeley file for citation.

\bibliography{BibWanSong}

\end{document}